\PassOptionsToPackage{numbers,sort&compress,square}{natbib}

\documentclass{article} 
\usepackage[preprint]{neurips_2025}

\usepackage[breaklinks=true]{hyperref}
\usepackage{url}
\usepackage{graphicx}
\usepackage{booktabs}
\usepackage{amssymb}
\usepackage{amsmath}
\usepackage{multirow}
\usepackage{multicol}

\usepackage{subcaption}

\usepackage{makecell}
\usepackage{pgffor}

\title{Ticket-Bench: A Kickoff for Multilingual and Regionalized Agent Evaluation}

\author{%
 Thales Sales Almeida\\
 Institute of Computing (IC)\\
 State University of Campinas \\
   Maritaca AI \\
   Tropic AI \\
  \And
  João Guilherme Alves Santos \\
  Institute of Computing (IC)\\
 State University of Campinas \\
    Tropic AI \\
  \AND
  Thiago Laitz \\
  Institute of Computing (IC)\\
 State University of Campinas \\
 Maritaca AI \\
    Tropic AI \\
  \And
  Giovana Kerche Bonás \\
Institute of Computing (IC)\\
 State University of Campinas \\
   Maritaca AI \\
    Tropic AI
}

\begin{document}

\maketitle

\begin{abstract}

Large language models (LLMs) are increasingly deployed as task-oriented agents, where success depends on their ability to generate accurate function calls under realistic, multilingual conditions. However, existing agent evaluations largely overlook cultural and linguistic diversity, often relying on monolingual or naïvely translated benchmarks. We introduce Ticket-Bench, a benchmark for multilingual agent evaluation in task-oriented scenarios. Ticket-Bench simulates the domain of soccer ticket purchases across six major languages—Portuguese, English, Spanish, German, Italian, and French—using localized teams, cities, and user profiles to provide a higher level of realism. We evaluate a wide range of commercial and open-source LLMs, measuring function-calling accuracy and consistency across languages. Results show that reasoning-oriented models (e.g., GPT-5, Qwen3-235B) dominate performance but still exhibit notable cross-lingual disparities. These findings underscore the need for culturally aware, multilingual benchmarks to guide the development of robust LLM agents.



\end{abstract}

\section{Introduction}

Large language models (LLMs) have quickly evolved from mere text generators to agents capable of orchestrating real-world actions through function-calling and tool use~\cite{patil2024gorilla, schick2023toolformer}. This paradigm shift has fueled the adoption of LLMs in a wide array of digital assistants and task automation platforms, where interpreting user requests and triggering appropriate actions is essential~\cite{agent_survey1, agent_survey2}.

A critical gap in current research is the absence of multilingual, culturally aware benchmarks for evaluating function-calling. Existing evaluations of tool use and agent performance cover important ground but are predominantly English-centric~\cite{eval_agent_survey, patilberkeley, castillo2024beyond, tau2bench}. Related efforts on general task completion and information retrieval extend to multiple languages~\cite{huang2025benchmax, multilingual_rag}, yet they often depend on monolingual or simply translated datasets. In real deployments, users converse with assistants in many languages and reference region-specific entities that shape model interactions and may influence how well a model executes function calls. Without benchmarks that reflect this linguistic and cultural localization, we cannot reliably assess—or improve—models’ ability to plan and fulfill real-world tasks across different regions.

To address this gap, we introduce Ticket-Bench, a benchmark for evaluating LLM function-calling in the domain of purchasing soccer game tickets. Ticket-Bench features tasks in six major languages—Portuguese, English, Spanish, German, Italian, and French. We carefully localize user queries and context, adapting city names, team names, and contextual nuances to each language and region. This approach ensures that LLMs are tested not only for multilingual understanding but also for their ability to handle realistic scenarios. Ticket-Bench is available at \url{https://github.com/TropicAI-Research/Ticket-Bench}.

Ticket-Bench provides a wide range of scenarios, requiring LLMs to interpret nuanced constraints and user preferences when generating structured function calls to interact with the system. Our evaluations reveal challenges in some LLMs to interpret user intent and produce the expected actions robustly across all tested languages.

The main contributions of our paper are as follows:

\begin{itemize}
\item We introduce Ticket-Bench, a benchmark designed to evaluate LLM agent capabilities in ticket-purchasing scenarios, featuring over 1000 evaluation cases across six languages with contextually adapted environments.
\item Ticket-bench provides an LLM-free, programmatic evaluation that checks the final environment state (expected tickets booked, no unexpected bookings). To capture robustness, we report a pass\^\space3 consistency metric computed over multiple executions per query, rewarding models that solve tasks reliably, not just once.
\item We observe systematic, family-specific language asymmetries: no language is uniformly “easy” or “hard,” but some model families show notable strengths and deficits across certain languages, likely due to language imbalances in the model training data. 
\end{itemize}
\section{Related work}

\textbf{Multilingual disparities in LLMs.} Recent studies have shown that LLMs perform unevenly across regions, particularly in underrepresented countries and cultural contexts. For example, WorldBench \cite{moayeri2024worldbench} exposes gaps in factual recall tied to economic and geographic divides, TiEBe \cite{almeida2025tiebe} highlights inconsistencies in capturing temporally grounded events, and BLEnD \cite{myung2024blend} uncovers cultural and linguistic biases in everyday knowledge.

While multilingual disparities are explored in text-based tasks, it is unclear how they impact LLMs function-calling capabilities, where models must correctly interpret and execute structured calls across languages and contexts.

\textbf{English-Only Function Call Benchmarks.} Most function-calling benchmarks for LLMs emphasize interaction realism and dialogue robustness, simulating policy-constrained dialogues or API-driven tasks. However, these evaluations are largely monolingual and culturally neutral, leaving open how models adapt to multilingual, localized settings.

\textbf{$\tau^2$-Bench}~\cite{barres2025tau2} extends the original $\tau$-Bench \cite{yao2024tau} with dual-control environments, where both agent and simulated user act on a shared state, and provides analyses that separate reasoning from coordination errors. \textbf{ConFETTI} evaluates turn-level function-calling across 109 human-simulated conversations (313 user turns) and 86 APIs, testing goal changes, follow-ups, and chained calls \cite{alkhouli2025confetti}. Both benchmarks focus on turn-level function-calling, assessing how well LLMs manage multi-turn interactions, handle dynamic goals, and coordinate with simulated users in English-only settings.

\textbf{HammerBench} evaluates LLM tool usage in long-context mobile assistant scenarios, incorporating multi-step task execution, error recovery, and realistic API sequences \cite{wang2025hammerbench}. \textbf{StableToolBench} focuses on robustness and reliability across diverse APIs, emphasizing consistency in multi-step interactions and using a virtual API server for stable evaluation \cite{guo2024stabletoolbench}. These benchmarks assess interactive, multi-step tool usage, testing how LLMs maintain context, execute workflows, and recover from errors.

\textbf{BigCodeBench} evaluates LLM code generation across multiple programming languages and frameworks \cite{zhuo2024bigcodebench}. \textbf{BFCL} focuses on chain-of-thought reasoning and problem-solving without interactive environment control \cite{patilberkeley}. \textbf{API-Bank} provides a collection of APIs and tasks for evaluation, focusing on single-task multi-step API usage; it does not involve multi-turn dialogue or agent-like decision making. These benchmarks share a focus on evaluating LLMs’ function-calling capabilities in single-task, multi-step scenarios, reasoning, and other aspects, but remain limited to only english.

\textbf{Multilingual Function-Calling Studies.} Recent work has begun to explore how multilingual contexts affect LLMs’ function-calling capabilities. For instance, \textbf{BenchMAX}~\cite{huang2025benchmax} introduced a multilingual evaluation suite with a Tool Use track, assessing models’ ability to invoke correct functions across multiple languages through simple translation of the nexus~\cite{nexusraven} dataset. Similarly, \textbf{ACEBench}~\cite{chen2025acebench} extends function-calling evaluation to both English and Chinese, providing insights into cross-lingual performance.


However, existing multilingual function-calling efforts remain limited in scope and experimental control. Most cover only a small set of languages or rely on direct translations, and do not cover the localization of relevant entities during evaluation. Furthermore, many evaluations further depend on LLM-as-judge or turn-level signals instead of verifiable end-state outcomes.

\textbf{TicketBench} provides a simulated, culturally grounded environment across six languages, with synchronized schedules, localized user profiles, and aligned question templates to ensure comparability. Models interact through a fixed set of fully translated functions with standardized inputs and outputs, allowing evaluation of reasoning and execution in multi-step function calls while isolating language-specific ambiguities. This design enables a more systematic and robust multilingual assessment than prior benchmarks.

Table~\ref{tab:benchmarks_methodology} summarizes the key characteristics of the benchmarks reviewed alongside TicketBench. \textit{Languages} indicates the number of languages supported by the benchmark; \textit{Regional Adaptation} reflects whether datasets incorporate localized attributes or cultural context; \textit{Interactiveness} denotes whether the model’s outputs dynamically influence the environment; \textit{Multi-Step} specifies whether tasks require sequential or dependent operations to fulfill the task; \textit{LLM-Free Evaluation} indicates whether correctness can be assessed without relying on another LLM as a judge; and \textit{System Focus} identifies benchmarks designed to evaluate full agent pipelines or realistic system workflows rather than isolated function calls. 

\begin{table*}[h]
\centering
\resizebox{\textwidth}{!}{%
\begin{tabular}{lcccccc}
\toprule
\textbf{Benchmark} & 
\makecell{\textbf{Languages}} &
\makecell{\textbf{Regional} \\ \textbf{Adaptation}} &
\makecell{\textbf{Interactiveness}} &
\makecell{\textbf{Multi-Step}} &
\makecell{\textbf{LLM-Free} \\ \textbf{Evaluation}} &
\makecell{\textbf{System} \\ \textbf{Focus}} \\
\midrule
$\tau^2$-Bench       & 1  & $\times$ & $\checkmark$ & $\times$     & $\checkmark$ & $\checkmark$ \\
ConFETTI            & 1  & $\times$ & $\checkmark$ & $\times$     & $\times$     & $\times$     \\
HammerBench         & 1  & $\times$ & $\times$ & $\checkmark$ & $\times$     & $\times$     \\
StableToolBench     & 1  & $\times$ & $\times$     & $\checkmark$ & $\times$     & $\times$     \\
BigCodeBench        & 1  & $\times$ & $\times$     & $\checkmark$ & $\checkmark$ & $\times$     \\
BFCL                & 1  & $\times$ & $\times$     & $\checkmark$ & $\checkmark$ & $\times$     \\
API-Bank            & 1  & $\times$ & $\times$     & $\checkmark$ & $\checkmark$ & $\times$     \\
ACEBench            & 2  & $\times$ & $\checkmark$ & $\checkmark$ & $\checkmark$ & $\checkmark$     \\
BenchMAX            & 17 & $\times$ & $\times$     & $\checkmark$ & $\checkmark$ & $\times$     \\
\midrule
\textbf{Ticket-Bench} & \textbf{6} & $\checkmark$ & $\checkmark$ & $\checkmark$ & $\checkmark$ & $\checkmark$ \\
\bottomrule
\end{tabular}
}
\caption{Comparison of function-calling benchmarks, highlighting multilingual coverage, regional adaptation, interactiveness, multi-step evaluation, LLM-free assessment, and system focus.}
\label{tab:benchmarks_methodology}
\end{table*}

\section{Methodology}

\subsection{Environment and Entities}

To evaluate function-calling capabilities in multilingual scenarios, we constructed a simulated ticket-purchasing environment with three main components.

\textbf{Users} are defined by a culturally appropriate name (sampled from common names in the target country), a virtual account balance, and a preferred soccer team. These attributes introduce personal constraints—such as affordability and team preference—into the simulation. For each language, we generate 20 users, ensuring that no two share the same preferred team.

\textbf{Game schedules} form the core set of events. Each schedule represents a full league season, with games specifying the home team, away team, city, stadium, ticket price, and date. All entity names (teams, cities, stadiums) are localized to the target language and region. We simulate two types of schedules per language: one where each of the 20 teams plays every other team once, and another where they play twice, producing a total of 380 matches—the same number found in most professional leagues that inspired our setup.

\textbf{Leaderboards} capture historical league performance, enabling queries that depend on past results. For each season, they record statistics for each team, including points, wins, draws, losses, goals scored, and goals conceded. To generate these tables, we synthetically assign goals to each match while enforcing consistency, ensuring the resulting distributions resemble realistic league outcomes.

\begin{table}[h]
\centering
\caption{Constraint coverage across the 17 templates. A checkmark (\checkmark) indicates that the constraint is present in the template.}
\label{tab:templates_constraints}
\resizebox{\textwidth}{!}{%
\begin{tabular}{@{}clccccc@{}}
\toprule
ID & Template (abridged) & Semester & Weekday & Price & Location & Leaderboard \\ \midrule
1  & Next \{user\_team\} game I can afford                 &            &         &       &         &            \\
2  & Next game of my team I can afford                     &            &         &       &         &            \\
3  & Next game I can afford, first semester                & \checkmark &         &       &         &            \\
4  & Next game I can afford, not on weekend                &            & \checkmark &       &         &            \\
5  & Cheapest game this year                               &            &         & \checkmark &         &            \\
6  & Next game in \{location\}                             &            &         &       & \checkmark &            \\
7  & Next game vs team with $>$60 points in \{year\}       &            &         &       &         & \checkmark \\
8  & Next game, second semester, midweek                   & \checkmark & \checkmark &       &         &            \\
9  & Most expensive game I can afford, not weekend         &            & \checkmark & \checkmark &         &            \\
10 & Cheapest game in \{location\}                         &            &         & \checkmark & \checkmark &            \\
11 & Next game in \{location\}, vs top 8 teams of \{year\} &            &         &       & \checkmark & \checkmark \\
12 & Cheapest game, second semester, midweek               & \checkmark & \checkmark & \checkmark &         &            \\
13 & Most expensive game I can afford, not weekend, in \{location\} &  & \checkmark & \checkmark & \checkmark &            \\
14 & Cheapest game in \{location\}, vs team with $>$20 goals \{year\} & &         & \checkmark & \checkmark & \checkmark \\
15 & Most expensive game I can afford, 2nd semester, midweek, in \{location\} & \checkmark & \checkmark & \checkmark & \checkmark & \\
16 & Cheapest game I can afford, not weekend, in \{location\}, vs $>$20 goals \{year\} &  & \checkmark & \checkmark & \checkmark & \checkmark \\
17 & Most expensive game, 2nd semester, not weekend, in \{location\}, vs top 3 of \{year1\}/\{year2\} & \checkmark & \checkmark & \checkmark & \checkmark & \checkmark \\ \bottomrule
\end{tabular}
}
\end{table}

\subsection{Question Templates}

We define 17 question templates representing distinct ticket-purchasing scenarios. Each template combines constraints drawn from five categories:

\begin{itemize}
    \item \textbf{Semester:} restricts the date to the first or second semester.
    \item \textbf{Weekday:} restricts the day of the week (e.g., avoids weekends or selects midweek days).
    \item \textbf{Price:} selects the cheapest or most expensive game that satisfies other constraints.
    \item \textbf{Location:} restricts the game to a specific city.
    \item \textbf{Leaderboard:} adds conditions based on past results (e.g., top $n$ teams, teams with more than $x$ goals or points).
\end{itemize}



Table~\ref{tab:templates_constraints} summarizes the 17 templates and the constraints each one involves. The actual templates can be found in the Appendix \ref{sec:template_list}.

For each question template, we instantiate ten unique queries by varying user profiles, game schedules, and constraint specifications. This results in 170 queries per language and a total of 1,020 queries across the six languages in TicketBench.

To capture a wider range of scenarios, we design a subset of queries for which no valid booking exists. Specifically, 15\% of the questions are constructed so that no game in the schedule satisfies the stated constraints. This guarantees that models are not only evaluated on their ability to find valid matches, but also on their capacity to correctly detect when no solution exists.

\subsection{Available Functions}

Models are provided with a fixed set of five callable functions to solve the user query. Each function exposes a simple interface with clearly defined inputs and outputs, ensuring consistency across languages. 

\textbf{Get User Info} retrieves the user profile, including the user’s name, preferred team, and current account balance. This function allows models to check affordability constraints and align ticket choices with the user preferences. 

\textbf{List Games} returns a paginated list of games, with a maximum of 10 entries per page. The function accepts optional filters on fields such as location and team, and also supports ordering (e.g., by date or by price). Each game is represented with its identifier, teams, city, stadium, ticket price, and scheduled date. This function is central for exploring the search space of possible tickets. 

\textbf{Buy Game Ticket} finalizes the decision process by purchasing a ticket for a given game identifier. Successful execution updates the environment state by reducing the user’s balance and marking the corresponding game as booked.

\textbf{Get Leaderboard} provides access to historical league performance. It returns, for a specified year, a table of per-team statistics including points, wins, draws, losses, goals scored, and goals conceded. This function enables queries that depend on conditions such as “top $n$ teams” or “teams with more than $x$ goals.” 

\textbf{Get Weekday from Date} returns the day of the week corresponding to a given date string in the format \texttt{YYYY-MM-DD}. This function supports constraints involving weekends or midweek games.

\subsection{Localization}

\begin{figure}
    \centering
    \includegraphics[width=1\linewidth]{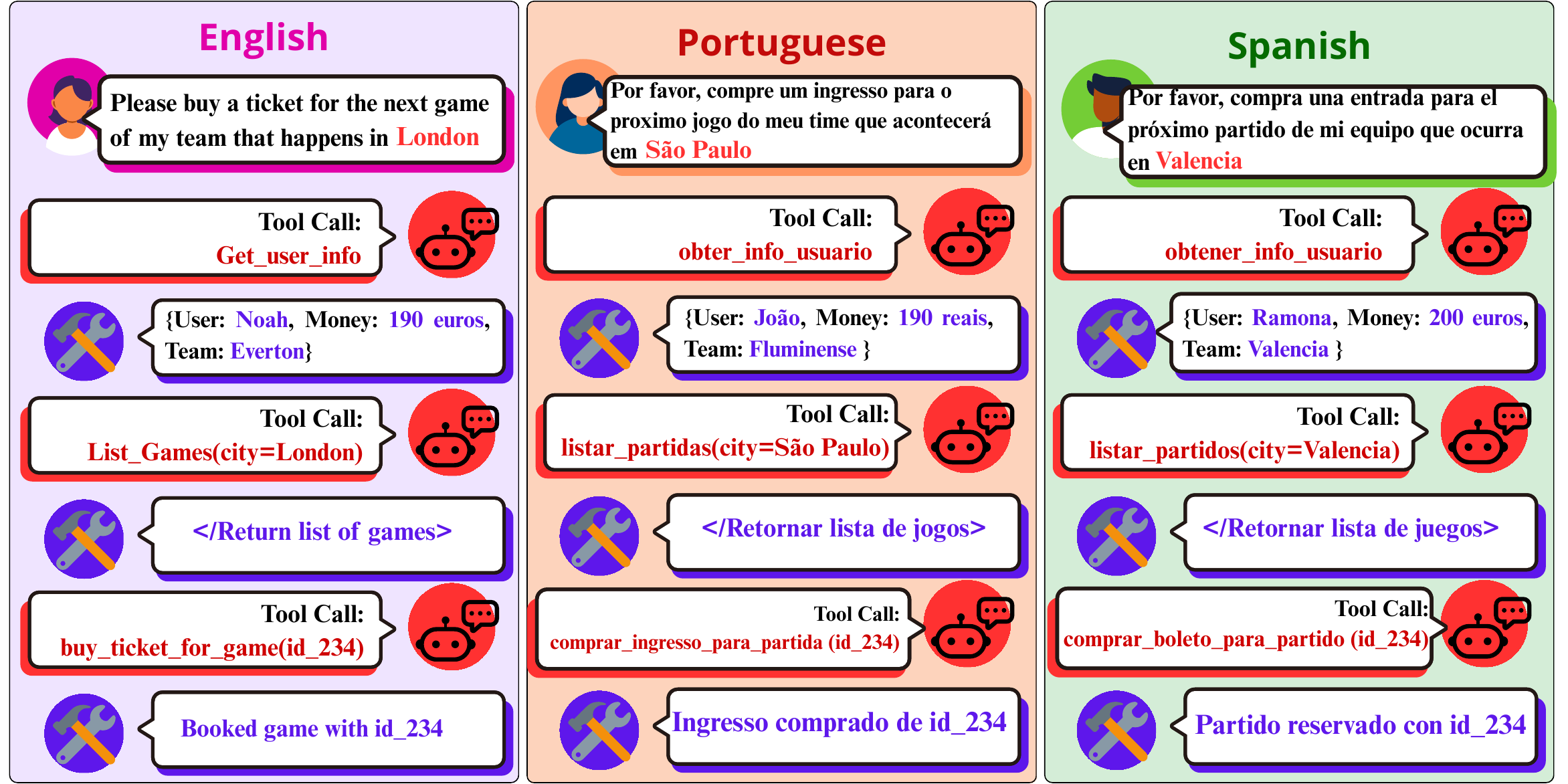}
    \caption{Example of Ticket Bench question localization}
    \label{fig:example_flow}
\end{figure}

We instantiate the environment in six languages, each aligned with a major national soccer league: the Brasileirão in Portuguese (Brazil), Ligue 1 in French (France), the Bundesliga in German (Germany), La Liga in Spanish (Spain), Serie A in Italian (Italy), and the Premier League in English (United Kingdom). For every language, team names are sourced from the official rosters of the respective leagues, and their home cities are used as the localized set of cities. User names are sampled from the most frequent names in each country to ensure cultural plausibility and naturalness in the generated scenarios.

To further control cross-linguistic comparability, we manually translate all question templates, along with the function names and their descriptions that are exposed to the model during execution. An illustration of the interaction flow is provided in Figure~\ref{fig:example_flow}. We enforce consistent constraints across languages and synchronize league schedules so that the distribution of games remains equivalent. This guarantees that differences in model performance during evaluation can be attributed to the agent's language-specific capabilities.

\subsection{Evaluation}



A query is considered correct if the resulting environment state after the LLM execution matches the annotated state, that is, all expected games are booked and no unexpected games are booked. This means our evaluation is not dependent on LLMs.

Our main metric, \textbf{pass\^\space k}, is adapted from code-generation benchmarks and estimates the probability that a model would succeed k independent attempts in the task. For each query $i$, let $c_i$ denote the number of correct executions across M runs. The empirical probability of success for each query is
\[
    p_i = \left(\frac{c_i}{M}\right)^k,
\]
And the overall score is given by
\[
    \text{pass}\space\hat{}\space \text{3} = \frac{1}{N} \sum_{i=1}^{N} p_i.
\]
This formulation rewards consistency across runs: queries that are solved correctly in multiple attempts contribute more than those solved only once.  We compute metrics separately for each target language and in aggregate over the full multilingual dataset. 

For this study, we set M=3 and K=3. We choose to run each model only 3 times for budget reasons, as some of the most expensive reasoning models can be notably expensive to run, GPT-5, for example, costs around \$70 USD to run 3 times in all of Ticket Bench.

\begin{table}[]
\centering
\caption{pass\^\space3 results for tested models in Ticket Bench. Results are displayed in each language covered in Ticket bench, and considering all questions in the benchmark.}
\label{tab:main_table}
\begin{tabular}{@{}l|ccccccc@{}}
\toprule
\multirow{2}{*}{model} & \multicolumn{7}{c}{Pass\^\space3}                        \\ \cmidrule(l){2-8} 
                       & en   & es   & fr   & it   & de   & pt   & overall \\ \midrule
GPT-5                  & 0.92 & 0.93 & 0.92 & 0.92 & 0.92 & 0.87 & 0.91    \\
GPT-5 Mini             & 0.91 & 0.91 & 0.89 & 0.90 & 0.89 & 0.86 & 0.89    \\
Qwen3-235B-A22B~\cite{yang2025qwen3}        & 0.88 & 0.91 & 0.90 & 0.85 & 0.88 & 0.86 & 0.88    \\
GPT-5 Nano             & 0.71 & 0.78 & 0.83 & 0.69 & 0.73 & 0.74 & 0.75    \\
GPT-OSS-120B~\cite{agarwal2025gpt}           & 0.73 & 0.76 & 0.72 & 0.70 & 0.73 & 0.67 & 0.72    \\
GPT-4.1                & 0.74 & 0.68 & 0.75 & 0.62 & 0.70 & 0.72 & 0.70    \\
Gemini-Pro 2.5~\cite{comanici2025gemini}                & 0.85 & 0.54 & 0.76 & 0.48 & 0.59 & 0.6 & 0.63    \\
Gemini-Flash 2.5~\cite{comanici2025gemini}                & 0.72 & 0.52 & 0.64 & 0.37 & 0.43 & 0.45 & 0.52    \\

Qwen3-32B~\cite{yang2025qwen3}              & 0.40 & 0.55 & 0.51 & 0.56 & 0.55 & 0.56 & 0.52    \\
GPT-4.1 Mini           & 0.48 & 0.59 & 0.49 & 0.52 & 0.54 & 0.48 & 0.52    \\
Qwen3-14B~\cite{yang2025qwen3}              & 0.34 & 0.46 & 0.40 & 0.38 & 0.44 & 0.45 & 0.41    \\
Qwen2.5-72B-Instruct~\cite{qwen25}   & 0.25 & 0.34 & 0.47 & 0.30 & 0.42 & 0.48 & 0.38    \\
Qwen2.5-32B-Instruct~\cite{qwen25}   & 0.28 & 0.30 & 0.37 & 0.25 & 0.35 & 0.43 & 0.33    \\
Qwen3-30B-A3B~\cite{yang2025qwen3}          & 0.24 & 0.34 & 0.36 & 0.31 & 0.38 & 0.35 & 0.33    \\
Sabia-3.1~\cite{sabia3}              & 0.30 & 0.24 & 0.27 & 0.21 & 0.29 & 0.30 & 0.27    \\
xLAM-2-32b-fc-r~\cite{xlam}        & 0.22 & 0.24 & 0.21 & 0.27 & 0.30 & 0.31 & 0.26    \\
Qwen3-8B~\cite{yang2025qwen3}               & 0.21 & 0.28 & 0.26 & 0.24 & 0.33 & 0.28 & 0.26    \\
GPT-OSS-20B~\cite{agarwal2025gpt}            & 0.29 & 0.27 & 0.21 & 0.14 & 0.29 & 0.31 & 0.25    \\
Qwen3-4B~\cite{yang2025qwen3}               & 0.20 & 0.25 & 0.22 & 0.22 & 0.27 & 0.22 & 0.23    \\
GPT-4.1 Nano           & 0.20 & 0.21 & 0.21 & 0.16 & 0.19 & 0.18 & 0.19    \\
Qwen2.5-14B-Instruct~\cite{qwen25}   & 0.16 & 0.18 & 0.22 & 0.12 & 0.11 & 0.25 & 0.17    \\
Qwen2.5-7B-Instruct~\cite{qwen25}    & 0.14 & 0.14 & 0.18 & 0.09 & 0.13 & 0.12 & 0.13    \\
Qwen2.5-3B-Instruct~\cite{qwen25}    & 0.10 & 0.12 & 0.12 & 0.08 & 0.11 & 0.10 & 0.11    \\
Llama-xLAM-2-8b-fc-r~\cite{xlam}   & 0.08 & 0.08 & 0.12 & 0.10 & 0.14 & 0.06 & 0.10    \\
xLAM-2-3b-fc-r~\cite{xlam}         & 0.03 & 0.11 & 0.07 & 0.05 & 0.06 & 0.05 & 0.06    \\ \bottomrule
\end{tabular}
\end{table}

\section{Results}

\subsection{Overall Performance}

Table~\ref{tab:main_table} reports pass\^\space3 results across all tested models, broken down by language and overall performance.

The five best-performing systems—GPT-5 (0.91), GPT-5 Mini (0.89), Qwen3-235B~\cite{yang2025qwen3} (0.88), GPT-5 Nano (0.75), and GPT-OSS-120B~\cite{agarwal2025gpt} (0.72)—all belong to the category of reasoning models. These systems are designed to allocate more inference cycles per query, often at higher computational cost. Their dominance suggests that reasoning-optimized architectures significantly improve multilingual function-calling.

Outside this cluster, accuracy starts to drop. GPT-4.1 (0.70 overall) remains a strong non-reasoning baseline, but its smaller variant GPT-4.1 Mini (0.52) falls behind, and GPT-4.1 Nano (0.19) shows a very lackluster performance. Qwen3-32B and Qwen3-14B show moderate performance (0.52 and 0.41), but the majority of Qwen2.5 models and smaller Qwen3 variants remain below 0.40 overall. This demonstrates that while instruction tuning improves usability, it does not provide the robustness required for complex multilingual function-calling.

An interesting trend emerges in the xLAM models~\cite{xlam}, which were fine-tuned for function-calling: their performance is consistently worse than their base Qwen2.5 models. For example, xLAM-2-32B-fc-r (0.26) underperforms Qwen2.5-32B-Instruct (0.33), and xLAM-2-3B-fc-r (0.06) falls behind Qwen2.5-3B-Instruct (0.11). This suggests that the specialized fine-tuning applied to xLAM may have improved capabilities in the target task but negatively impacted generalization across languages and tasks.




\subsection{Scaling Trends}

\begin{figure}[htbp]
    \centering

    \begin{subfigure}[b]{0.48\textwidth}
        \centering
        \includegraphics[width=\textwidth]{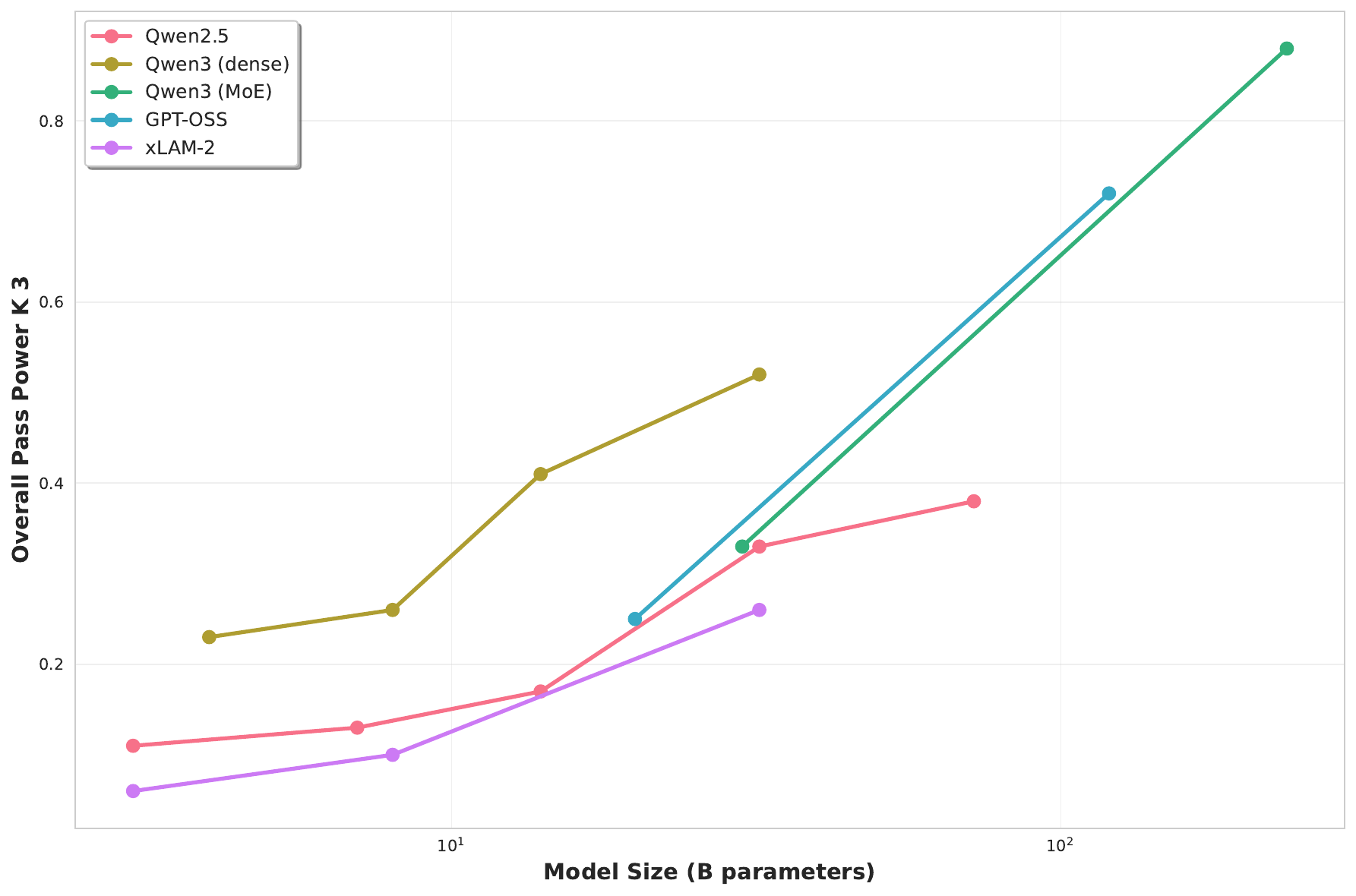}
        \caption{Scaling Law of various LLMs families on Ticket-Bench.}
        \label{fig:scaling_pass3}
    \end{subfigure}
    \hfill
    \begin{subfigure}[b]{0.48\textwidth}
        \centering
        \includegraphics[width=\textwidth]{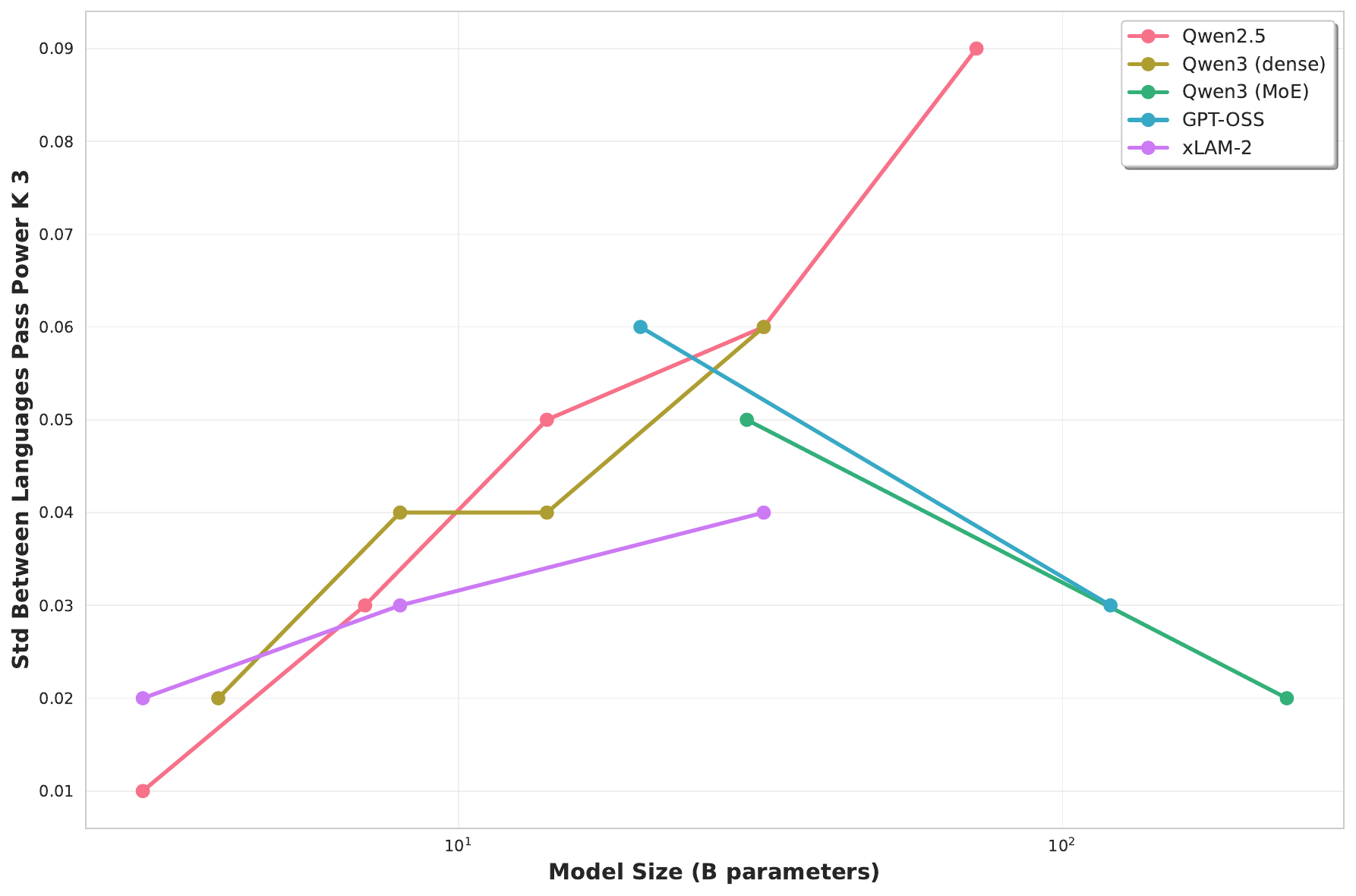}
        \caption{Standard deviation between different languages of various LLM families.}
        \label{fig:scaling_std}
    \end{subfigure}

    \caption{Scaling tendencies of open source models in Ticket Bench.}
    \label{fig:poetav2_type_and_subcategory_dist}
\end{figure}

Figures~\ref{fig:scaling_pass3} and~\ref{fig:scaling_std} provide additional perspective by analyzing scaling behavior across model families.  


Figure~\ref{fig:scaling_pass3} illustrates the scaling behavior of different LLM families on Ticket Bench. The results follow a clear scaling law: as model size increases, accuracy steadily improves across most families. However, the slope of this improvement differs significantly. The Qwen3 (MoE) family shows the steepest growth, reaching competitive performance at large parameter counts. GPT-OSS models also follow a near-linear scaling trend, reinforcing the value of additional capacity. In contrast, the Qwen2.5 and xLAM-2 families scale more slowly, plateauing at considerably lower performance levels. This suggests that scaling alone is not sufficient; architectural and training choices (e.g., reasoning optimization) strongly mediate gains from larger parameter budgets.


Figure~\ref{fig:scaling_std} reports the standard deviation of accuracy across languages for each family. Larger models generally exhibit greater cross-lingual consistency, as evidenced by the declining variance in GPT-OSS and Qwen3 (MoE) models at the highest scales. By contrast, Qwen2.5, Qwen3 (dense) and XLAM-2 models show increasing variance as parameters grow, indicating uneven improvements across languages.  

\begin{figure}
    \centering
    \includegraphics[width=0.85\linewidth]{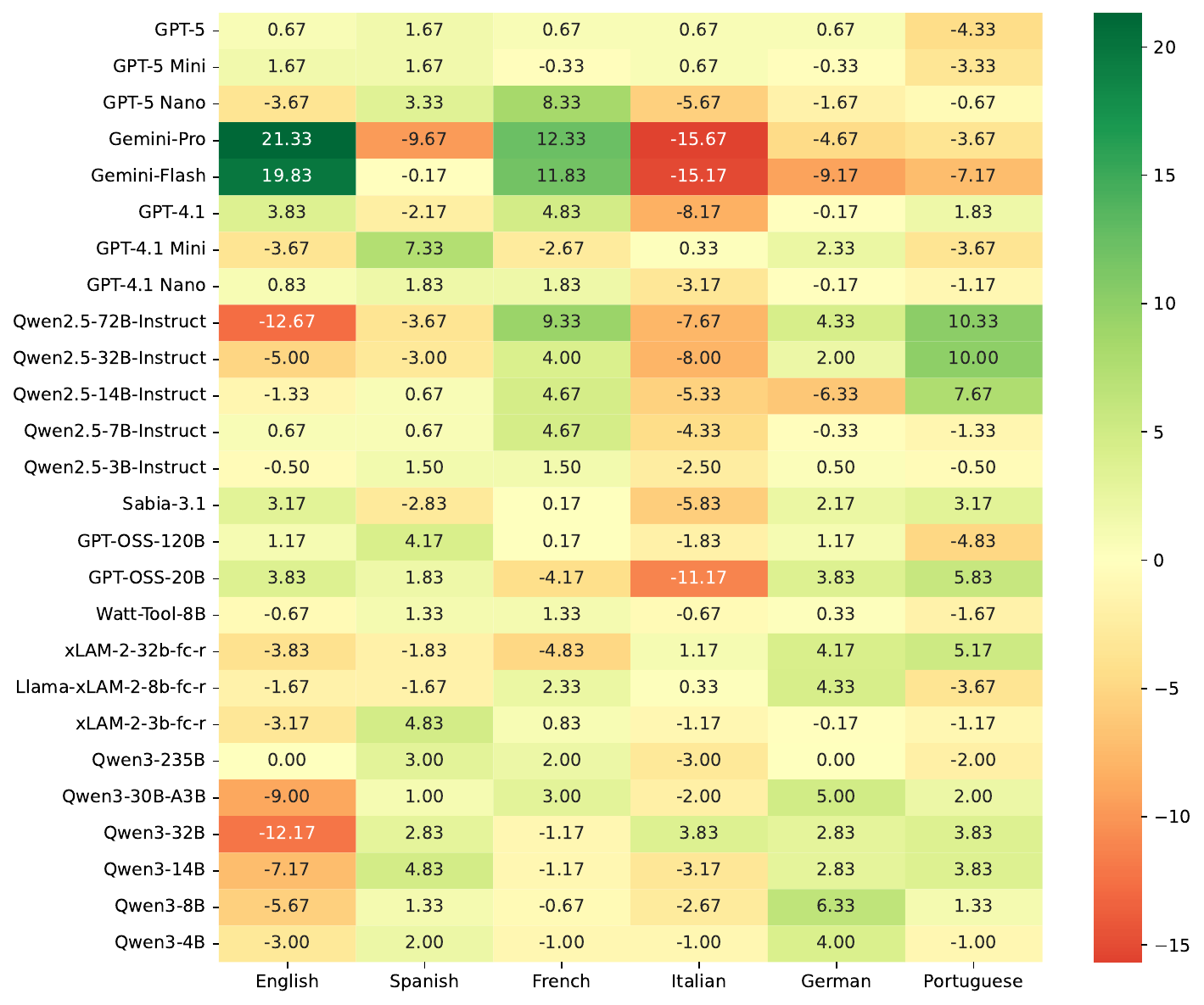}
    \caption{Heatmap showing the difference between each model per-language pass\^\space3 performance and its own mean pass\^\space3 performance among all languages. Models of the same family are displayed together.}
    \label{fig:heatmap_langvar}
\end{figure}

\subsection{Cross-lingual Variation}

Figure~\ref{fig:heatmap_langvar} analyzes the relative performance of each model across languages by subtracting the model’s overall average performance from its per-language score. Positive values indicate languages where the model performs better than its own mean, while negative values highlight languages where the model performs worse than it's average performance.

\textbf{No “easy” or “hard” language across the board.}. No single language consistently depresses scores across all systems. Instead, each language interacts differently with different families. For example, Qwen2.5-72B and Qwen2.5-32B achieve strong gains in Portuguese but show sharp deficits in English and Italian, while GPT-4.1 performs well in French yet struggles in Italian.

\textbf{Family-specific asymmetries}. Certain families show systematic biases. Qwen2.5 instruct models tend to favor French and Portuguese but lose accuracy in English and Italian. Qwen3 models also display a relative drop in English while improving in German and Spanish. Most surprisingly, both Gemini models display a disproportionate increase in English performance, with a secondary gain in French, but weaker results in all other languages..By contrast, the largest GPT-5 models maintain more balanced cross-lingual results, with the main exception being Portuguese, where deviations are more pronounced. These family-specific patterns are likely a reflection of the training data distribution used in each family.

\textbf{The best performing models are more robust.}. The strongest models overall—GPT-5, GPT-5 Mini, and Qwen3-235B—are also the most consistent across languages. Their performance remains close to their own mean, indicating greater robustness. Nevertheless, even these models exhibit differences of at least five points between their best- and worst-performing languages, showing that cross-lingual performance remains an open challenge.

Taken together, these findings show that multilingual function-calling remains uneven among most advanced models. While scale and reasoning can contribute to reducing variation, residual gaps across languages indicate that balanced and diverse multilingual training is still necessary.

\section{Conclusion}
In this work, we introduced Ticket-Bench, a benchmark designed to evaluate multilingual capabilities and variations in LLM agents. By simulating soccer ticket purchases across six major languages with localized teams, cities, and user profiles, Ticket-Bench provides a systematic and realistic framework for assessing LLM agent capabilities in multiple languages.

Our experiments reveal three central findings. First, reasoning-oriented models such as GPT-5 and Qwen3-235B show the most impressive performance. Second, scaling trends confirm that larger models generally achieve higher accuracy and more consistent results across languages, though families differ in their scaling efficiency. Third, cross-lingual variation remains a persistent challenge: no language is universally “easy” or “hard,” and model families exhibit distinct asymmetries, underscoring the influence of training data distributions.


Ticket-Bench highlights both the progress of state-of-the-art reasoning models and their limitations. We hope that this benchmark will serve as a foundation for future research, encouraging the design of models and training regimes that are not only more powerful but also more equitable and reliable across the diverse linguistic and cultural contexts in which they will ultimately be deployed.

\bibliography{preprint}
\bibliographystyle{plainnat}

\appendix

\section{Template List}
\label{sec:template_list}

This appendix shows the English version of all the templates used in Ticket Bench. Placeholders denoted between \{\} are dynamically filled when instantiating the templates.

\begin{quote}
\begin{enumerate}
    \item Please buy a ticket for the next \texttt{\{user\_team\}} game that I can afford.
    \item Please buy a ticket for the next game of my team that I can afford.
    \item Please buy a ticket for the next game of my team that I can afford, and that happens in the first semester of the year.
    \item Please buy a ticket for the next game of my team that I can afford and that is not on a weekend.
    \item Please buy a ticket for the cheapest game of my team that happens this year.
    \item Please buy a ticket for the next game of my team that happens in \texttt{\{location\}}.
    \item Please buy a ticket for the next game of my team that is against a team that scored more than 60 points in \texttt{\{year\}}.
    \item Please buy a ticket for the next game of my team that happens in the second semester of the year and that takes place in the middle of the week (Tuesday, Wednesday, or Thursday).
    \item Please buy a ticket for the most expensive game of my team that I can afford and that is not on a weekend.
    \item Please buy a ticket for the cheapest game of my team that is in \texttt{\{location\}}.
    \item Please buy a ticket for the next game of my team that happens in \texttt{\{location\}} and is against one of the top 8 teams of \texttt{\{year\}}.
    \item Please buy a ticket for the cheapest game of my team that happens in the second semester of the year and that takes place in the middle of the week (Tuesday, Wednesday, or Thursday).
    \item Please buy a ticket for the most expensive game of my team that I can afford and that is not on a weekend and that is in \texttt{\{location\}}.
    \item Please buy a ticket for the cheapest game of my team that is in \texttt{\{location\}} and is against a team that scored more than 20 goals in \texttt{\{year\}}.
    \item Please buy a ticket for the most expensive game of my team that I can afford and that happens in the second semester of the year, takes place in the middle of the week (Tuesday, Wednesday, or Thursday), and is in \texttt{\{location\}}.
    \item Please buy a ticket for the cheapest game of my team that I can afford, that is not on a weekend, is in \texttt{\{location\}}, and is against a team that scored more than 20 goals in \texttt{\{year\}}.
    \item Please buy a ticket for the most expensive game of my team that I can afford, that is in \texttt{\{location\}}, is against one of the top 3 teams of \texttt{\{year1\}} or \texttt{\{year2\}}, that is not on a weekend, and that happens in the second semester of the year.
\end{enumerate}
\end{quote}






\end{document}